\documentclass{article}

\PassOptionsToPackage{numbers,compress}{natbib}
\usepackage[preprint]{neurips_2026}

\usepackage[utf8]{inputenc}
\usepackage[T1]{fontenc}
\usepackage{hyperref}
\usepackage{url}
\usepackage{graphicx}
\usepackage{subcaption}
\usepackage{booktabs}
\usepackage{amsfonts}
\usepackage{nicefrac}
\usepackage{microtype}
\usepackage[table,xcdraw]{xcolor}
\usepackage{threeparttable}
\usepackage{multirow}
\usepackage{makecell}
\usepackage{wrapfig}
\usepackage{float}
\usepackage{amsmath}
\usepackage{amssymb}
\usepackage{mathtools}
\usepackage{amsthm}
\usepackage[capitalize,noabbrev]{cleveref}
\usepackage[disable,textsize=tiny]{todonotes}

\theoremstyle{plain}

\theoremstyle{definition}

\theoremstyle{remark}

\title{FlashEdit: Decoupling Speed, Structure, and Semantics for Precise Image Editing}

\author{
  Junyi Wu$^{1}$\thanks{Equal contribution}~,\enspace
  Zhiteng Li$^{1}$\footnotemark[1]~,\enspace
  Haotong Qin$^{2}$,\enspace
  Yulun Zhang$^{1}$\thanks{Corresponding author: Yulun Zhang, yulun100@gmail.com}~,\enspace
  Xiaokang Yang$^{1}$\enspace \\
  \textsuperscript{1}Shanghai Jiao Tong University\enspace
  \textsuperscript{2}ETH Z\"{u}rich\enspace
}

\begin{document}

\maketitle
\begin{abstract}
Text-guided image editing with diffusion models has achieved remarkable quality but often suffers from prohibitive latency. We introduce \textbf{FlashEdit}, a real-time localized image editing framework for the standard inversion-based editing setting. Its efficiency and precision stem from three key innovations: (1) a \textbf{Cycle-Consistent One-Step Inversion (COSI)} pipeline that encourages manifold-aligned one-step inversion through cycle consistency; (2) a \textbf{Background Shield (BG-Shield)} technique that improves preservation of non-edited regions via structural self-attention intervention; and (3) a \textbf{Sparsified Spatial Cross-Attention (SSCA)} mechanism that promotes precise edits by suppressing semantic leakage. Experiments on PIE-Bench demonstrate a strong preservation-efficiency trade-off, with edits completed in under 0.2 seconds and an over 150$\times$ speedup over DDIM-based multi-step editing. Our code will be made publicly available at \url{https://github.com/JunyiWuCode/FlashEdit}.
\end{abstract}

\begin{figure}[H]
  \centering
  \includegraphics[width=1.0\linewidth]{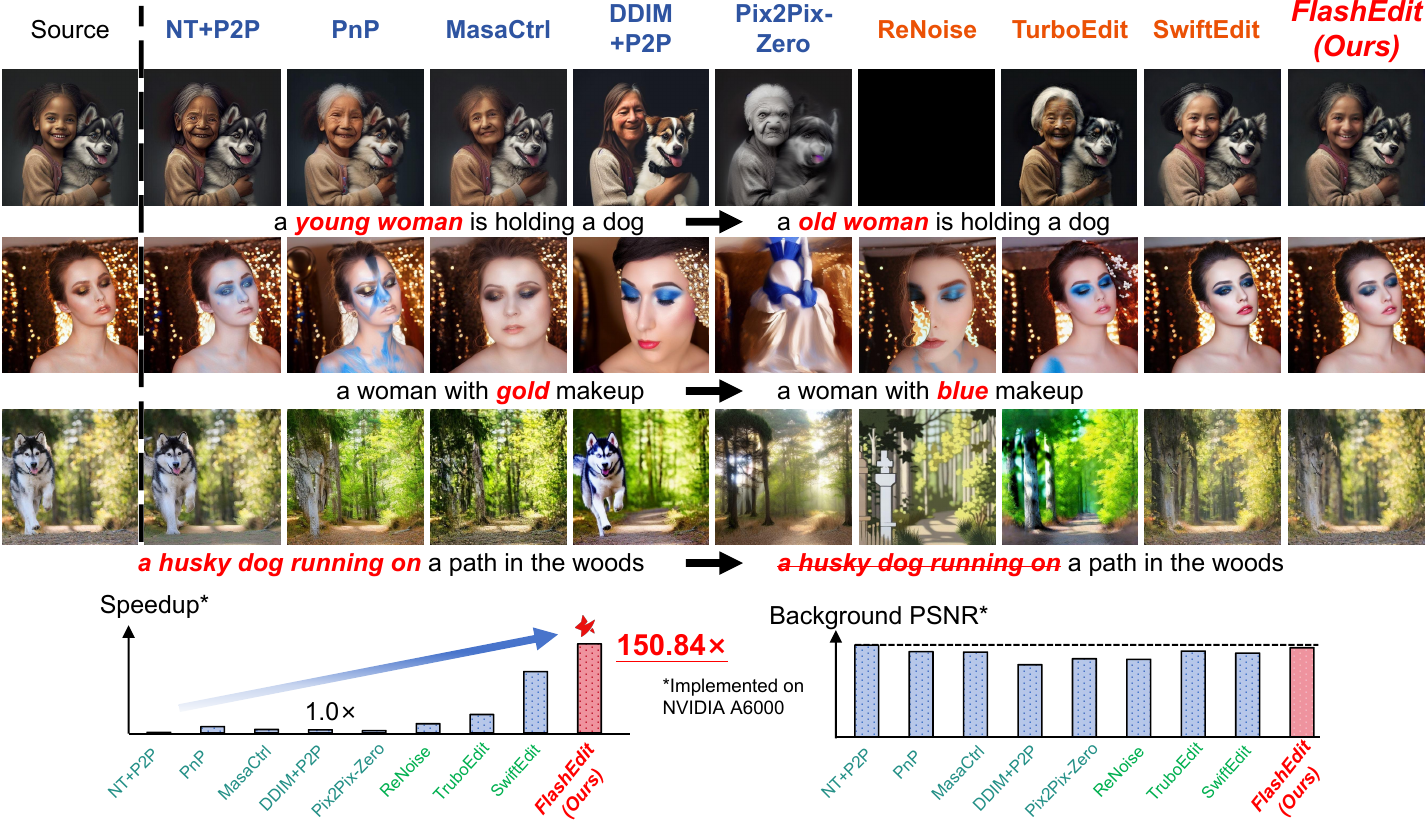}
  \caption{\textbf{Overview and qualitative examples of FlashEdit.} FlashEdit targets real-time localized text-guided editing by one-step inversion, preservation of regions outside the edit mask, and sparse semantic guidance.}
  \label{fig:overview}
\end{figure}
\section{Introduction}

Text-guided image editing with diffusion models~\cite{instructpix2pix, dong2023prompt} has demonstrated remarkable capabilities, allowing users to perform complex semantic modifications with high fidelity. The standard methodology is built upon a two-stage inversion-denoising pipeline: an initial inversion process maps a source image to its corresponding noise latent, which is then progressively denoised to generate the edited output according to a target prompt~\cite{directInversion,masactrl}. The objective is to achieve high fidelity in both content preservation and target prompt alignment, which often necessitates a computationally intensive, multi-step process.

Recent research has pursued several distinct strategies to improve accuracy and speed. To tackle the latency of the multi-step denoising process, methods based on model distillation have been proposed to enable editing in a faster way~\cite{turboedit}. These approaches must carefully address challenges such as mismatched noise statistics and insufficient editing strength that arise when adapting multi-step frameworks to fast samplers~\cite{mokady2023null, miyake2025negative}. To improve edit precision and prevent semantic leakage into the background, another category of work modifies the model's internal mechanisms, primarily by re-weighting or replacing attention maps to make the edit more spatially constrained~\cite{alphaedit, headrouter}. Recognizing that the final edit quality is highly dependent on the starting point, other approaches focus on refining the inversion technique itself~\cite{directInversion}. These methods aim to find a more accurate initial latent vector, with recent insights revealing that separating the objectives of content preservation and edit fidelity can yield significant performance gains and speedups~\cite{wang2025image}.

However, these existing methods approach speed and quality as a trade-off rather than as interconnected components of a singular, complex control problem. They offer partial solutions like accelerating the sampler at the cost of inversion fidelity, or preserving non-edited regions without addressing the precision of the masked edit region. This leaves room for a unified framework that jointly considers latency, preservation, and semantic control.

To address this multifaceted challenge, we introduce a novel editing methodology that establishes control at three progressively finer levels of granularity. At the foundational level, we tackle the macro-problem of \textbf{temporal control}. At the foundational level, we tackle the macro-problem of \textbf{manifold alignment}. We propose a \textbf{Cycle-Consistent One-Step Inversion (COSI)} pipeline, built upon a bijective loop strategy, which reduces the latency of prior iterative inversion while maintaining high reconstruction fidelity. With this temporal control established, we address the meso-level problem of \textbf{spatial control}. Our \textbf{Background Shield (BG-Shield)} mechanism improves structural integrity by intervening in the self-attention layers. It uses a memory of non-edited regions and queries from the editable region to separate edited and preserved features, improving spatial stability. Finally, with temporal and spatial controls in place, we target the micro-level problem of \textbf{semantic control}. We develop \textbf{Sparsified Spatial Cross-Attention (SSCA)}, a refinement of the cross-attention mechanism that prunes irrelevant text tokens pre-softmax, encouraging the edit to follow a focused semantic signal. Each component logically builds upon the last, forming a cohesive solution (Figure~\ref{fig:overview}). Our main contributions can be summarized as follows:
\begin{itemize}
    \item We propose a novel, multi-level methodology for image editing that cohesively integrates control over three distinct levels: the temporal latency of the pipeline, the spatial structure of the image, and the semantic content of the edit with an over $150\times$ speedup compared to prior multi-step methods.
    
    \item At the temporal and geometric level, we introduce the \textbf{Cycle-Consistent One-Step Inversion (COSI)} pipeline, which uses a cycle-consistency objective to encourage alignment between image and noise latents under the generator's learned manifold.
    
    \item At the spatial level, we propose \textbf{Background Shield (BG-Shield)}, a structural intervention in self-attention that uses memory caching and soft recomposition with attention feathering to improve preservation of regions outside the editable mask.
    
    \item At the semantic level, we develop \textbf{Sparsified Spatial Cross-Attention (SSCA)}, a cross-attention mechanism that performs pre-softmax token pruning. This provides the final layer of fine-grained control, eliminating attribute bleeding and enabling precise edits with complex text prompts.
\end{itemize}

\vspace{-5pt}
\begin{figure}[t]
\begin{center}
    \centering
    \includegraphics[width=1.0\linewidth]{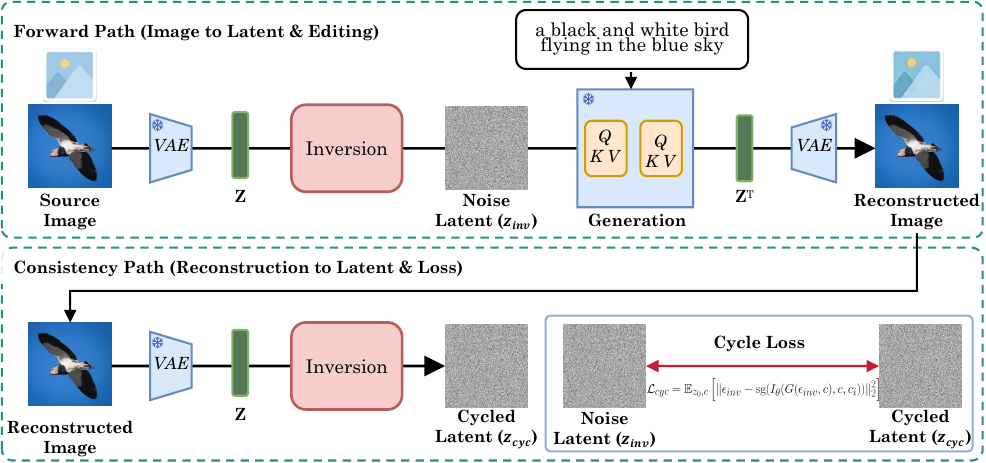}
    \vspace{-3mm}
    
\end{center}
\caption{\textbf{Overall pipeline of FlashEdit.} Given a source image, source prompt, target prompt, and edit mask, FlashEdit performs one-step inversion with COSI and applies localized editing with BG-Shield and SSCA to balance latency, preservation, and semantic alignment.}
   \label{fig:method1}
    \vspace{-5.5mm}
\end{figure}
%

\section{Related Works}
\subsection{Diffusion Models}
Recent advances in image synthesis have been largely driven by diffusion models~\cite{peebles2023scalable, kulikov2024flowedit}, which have become a leading paradigm for generating high-fidelity images from text. The core mechanism involves an iterative denoising process that progressively refines a random noise vector into a coherent image conditioned on a text prompt. A landmark contribution in this area is Stable Diffusion
~\cite{stableDiffusion}, a Latent Diffusion Model (LDM)~\cite{rombach2022high} that performs the computationally intensive denoising process in a lower-dimensional latent space, making the technology widely accessible. Parallel to this, Flow Matching models like Flux~\cite{flux2024} maps noise to an image via a more direct, straight-line trajectory, representing a different theoretical foundation for high-quality generative modeling.

To mitigate the high computational cost of these iterative models, various acceleration techniques have been proposed. Model quantization~\cite{li2024large,li2024fast,arbllm,pbs2p}, cache mechanism~\cite{specee,specdiff}, sparse attention~\cite{sparseAttn}, pruning~\cite{specprunevla,yan2025recalkv}, and distillation~\cite{hinton2015distilling,nguyen2025swiftedit} are general acceleration techniques for deep learning model. In diffusion models, specifically, one primary category is \textit{model quantization}~\cite{dvdquant}, which reduces memory footprint and computational load by converting full-precision model weights and activations into lower-bit representations. 
Post-Training Quantization (PTQ) is a particularly popular training-free approach in this domain.
Another category involves \textit{cache mechanisms}~\cite{liu2025reusing,deepcache,TaylorSeer2025}, which enhance inference efficiency by exploiting temporal redundancy. These methods reuse intermediate features computed at earlier denoising steps to avoid redundant calculations in later steps. While effective in isolation, recent work like QuantCache~\cite{quantcache} demonstrates a unified framework can yield greater gains.

\begin{figure}[t]
\begin{center}
    \centering
    \includegraphics[width=1\linewidth]{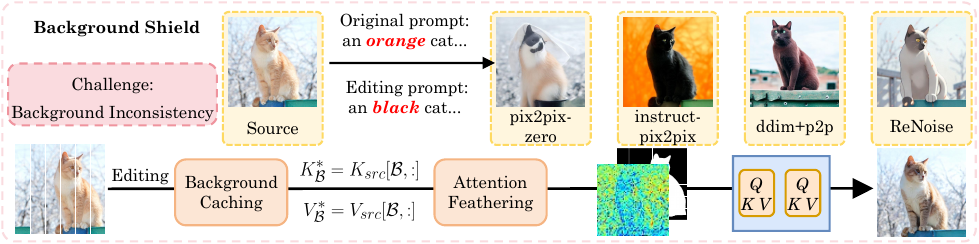}
    \vspace{-7mm}
\end{center}
\caption{\textbf{Cycle-Consistent One-Step Inversion (COSI).} COSI maps the source image to a one-step noise latent for reconstruction and re-inverts the reconstruction through a consistency path. The cycle loss encourages compatibility between the inversion network and the generator's one-step reconstruction trajectory.}

   \label{fig:method2}
\end{figure}

\subsection{Editing Models}
The task of editing real images with pre-trained generative models introduces the fundamental challenge of \textit{inversion}: finding a latent representation that can faithfully reconstruct a given source image. This problem was first extensively studied in the context of Generative Adversarial Networks (GAN) Inversion~\cite{wang2021HFGI,zhu2020domain,zhu2016generative}. In comparison, \textbf{DDIM Inversion}~\cite{song2020denoising} provides a deterministic method to find a corresponding noise latent for a source image. Once this latent is obtained, various editing mechanisms are employed during the denoising process to apply the desired changes. A prominent family of methods focuses on \textit{attention control}, where the cross-attention maps between text and image are manipulated. For example, to change a ``photo of a red car'' to a ``blue car,'' {Prompt-to-Prompt}~\cite{p2p} identifies the attention weights corresponding to the word ``red'' and replaces them with those for ``blue,'' preserving the attention for ``car'' and the background. Another powerful technique is \textit{feature injection}, exemplified by Plug-and-Play (PnP)~\cite{pnp}. To preserve the identity of a subject, PnP injects the self-attention features---which encode structure and appearance---from the source image's generation process into the edited one.
This helps keep facial identity consistent when changing a ``photo of a person'' to a ``photo of a person smiling.''
A third approach is \textit{mask-based editing}, where methods like {DiffEdit}~\cite{couairon2022diffedit} generate a mask indicating the region to be altered and then apply the denoising process only within that area. Despite these advances, cleanly separating the edited region from the preserved region remains challenging. Furthermore, most of these established techniques rely on computationally intensive multi-step inference, which severely hinders real-time interaction. While recent efforts have explored accelerating editing via model distillation~\cite{renoise,turboedit,nguyen2025swiftedit}, they often face a delicate balance between structural preservation and semantic alignment.

\section{Method}
\subsection{Cycle-Consistent One-Step Inversion}

\noindent\textbf{Challenge: Manifold Misalignment and Latent Drift.$\quad$}
Traditional inversion for multi-step diffusion relies on iterative refinement to push the noise latent into a region that accurately reconstructs the source image. However, in one-step generative frameworks, the solution space is narrow, and the generative trajectory is highly sensitive to the initial noise distribution. While distillation-based methods~\cite{nguyen2025swiftedit} attempt to map images to noise via a teacher's guidance, they can still exhibit \textit{manifold misalignment}. In one-step models, the generator $G$ characterizes a learned manifold $\mathcal{M}_G$ where the generative dynamics are significantly more compressed than multi-step counterparts. In such a compressed space, the model has less "self-correction" capacity than multi-step sampling. The predicted noise $\epsilon_{inv}$ from heuristic distillation may yield a visually plausible reconstruction $\hat{z}_0$ at the pixel level, but small mismatches can be amplified when the latent is perturbed by a target prompt $c_{tgt}$ for editing. Without subsequent denoising steps to buffer this error, it can manifest as \textit{latent drift}, including visible background flickering or structural changes.

\noindent\textbf{Motivation: From Mimicry to Bijective Manifold Anchoring.$\quad$}
Our key insight is that a stable inversion should not merely mimic a teacher's noise prediction, but should also be consistent with the generator $G$ used for one-step decoding. In a one-step setting, where the diffusion ODE is approximated by a direct mapping, the relationship between the image latent $z_0$ and the noise $\epsilon$ should ideally be close to an invertible mapping within the generative subspace. Without such a constraint, the inversion mapping can be under-constrained, sacrificing reconstructive fidelity or editability. We hypothesize that a \textbf{Cycle-Consistency Constraint} can regularize the inversion network $I_\theta$ toward a generator-consistent noise latent. This cycle-consistency objective encourages the inverted latent to remain compatible with the generator's one-step reconstruction trajectory, which we refer to as \textit{manifold alignment}. By making the inversion more aware of the generator's reconstruction behavior, COSI improves the stability of subsequent localized edits.

\noindent\textbf{Proposed Method: Cycle-Consistent Bijective Loop.$\quad$}
To resolve these issues, we propose \textbf{COSI}, which moves beyond simple distillation by integrating a self-supervised alignment objective. We define the inversion mapping as $\epsilon_{inv} = I_\theta(z_0, c, c_i)$, where $c$ is the text condition and $c_i$ is the visual adapter feature. While an initial foundational mapping is established using existing one-step training paradigms, our core contribution lies in reducing the manifold gap through the \textbf{Inversion Cycle}. Unlike previous works that treat the inversion network as a black-box regressor, COSI treats it as a partner to the generator and optimizes a closed-loop consistency objective.

For any given reconstruction $\hat{z}_0 = G(\epsilon_{inv}, c)$, the cycle-consistency principle dictates that feeding this reconstructed image back into the inversion network must recover the original noise $\epsilon_{inv}$. We optimize $I_\theta$ by minimizing the deviation between the initial predicted noise and the noise re-extracted from the reconstruction:
\begin{equation}
    \mathcal{L}_{cyc} = \mathbb{E}_{z_0, c} \left[ \left\| \epsilon_{inv} - \text{sg}(I_\theta(G(\epsilon_{inv}, c), c, c_i)) \right\|_2^2 \right],
\end{equation}
where $\text{sg}(\cdot)$ denotes the stop-gradient operation to guide the inversion network toward the generator's fixed output characteristics. Unlike traditional distillation losses that rely on an external teacher $\phi$, $\mathcal{L}_{cyc}$ provides a generator-aware signal tied to $G$. This objective encourages $I_\theta$ to learn an inverse mapping that is consistent with the one-step generator by verifying its own predictions through the generator's reconstruction path. By closing this loop, COSI provides a stable starting point that supports background preservation and localized semantic editing.

\begin{figure}[t]
\begin{center}
    \centering
    \includegraphics[width=1\linewidth]{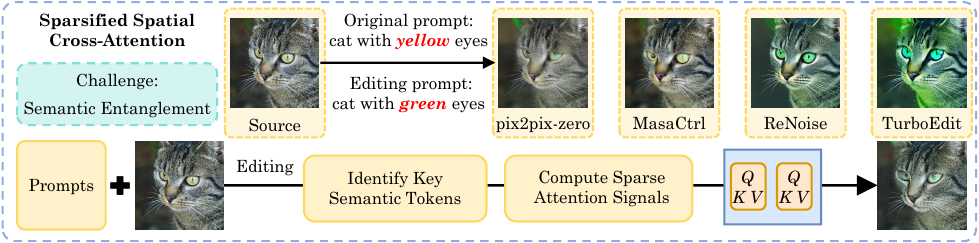}
    \vspace{-7mm}
    
\end{center}
\caption{\textbf{Sparsified Spatial Cross-Attention (SSCA).} Standard cross-attention may entangle edit-relevant and irrelevant tokens. SSCA selects the most relevant text tokens for the editable region and computes a sparse attention signal to reduce attribute leakage.}

   \label{fig:method3}
\end{figure}

\subsection{Background Shield}
\noindent\textbf{Challenge: Preserving Non-Edited Regions.$\quad$}
A critical challenge in localized image editing is preserving the regions outside the user-specified edit mask. We observe that even with precise masks, many methods can introduce visible changes outside the intended edit region. For instance, in Figure~\ref{fig:method2}, when performing a seemingly simple edit such as changing ``an orange cat'' to ``a black cat'', regions outside the mask may shift in color, lighting, or style. We identify the root cause of this instability as the inherent nature of the self-attention mechanism. As a global operator that computes all-to-all relationships between image tokens, it can allow the strong semantic signal from the masked edit region to propagate into preserved regions, undermining localized editing.

\noindent\textbf{Motivation.$\quad$}
Having identified the global nature of self-attention as a source of leakage outside the edit mask, our motivation is to move beyond merely scaling influences and propose a direct structural intervention. BG-Shield is designed for localized editing, where the mask defines the editable region and all unmasked pixels are treated as regions to preserve. Thus, the method does not assume that only semantic foreground objects can be edited: if a user wants to edit the sky, road, or another background area, that region is included in the editable mask. We introduce \textbf{Background Shield (BG-Shield)}, a method designed to improve consistency by recalling features for the unedited region from a cached memory.

\noindent\textbf{Proposed Method.$\quad$}
Shown in Figure~\ref{fig:method2}, BG-Shield operates as a two-pass mechanism within self-attention layers. Let $X \in \mathbb{R}^{S \times D}$ be the input feature sequence, and let a binary mask $M \in \{0, 1\}^S$ define the editable indices $\mathcal{F}$ and preserved indices $\mathcal{B}$. We keep the notation $\mathcal{F}$ and $\mathcal{B}$ for consistency with common foreground/background terminology, but they refer to mask-defined editable and preserved regions rather than fixed semantic categories.

 \noindent\textbf{Memory Caching for Preserved Regions.$\quad$} During a forward pass with the source prompt $c_{src}$, we compute the Key and Value matrices, $K_{src}, V_{src}$. We then extract and cache the key-value pairs outside the editable mask:
\begin{equation}
    K^*_{\mathcal{B}} = K_{src}[\mathcal{B}, :], \quad V^*_{\mathcal{B}} = V_{src}[\mathcal{B}, :].
\end{equation}
This cached memory, $(K^*_{\mathcal{B}}, V^*_{\mathcal{B}})$, serves as a high-fidelity record of the original state in the preserved region.

\begin{figure}[t]
\begin{center}
    \centering
    \includegraphics[width=1\linewidth]{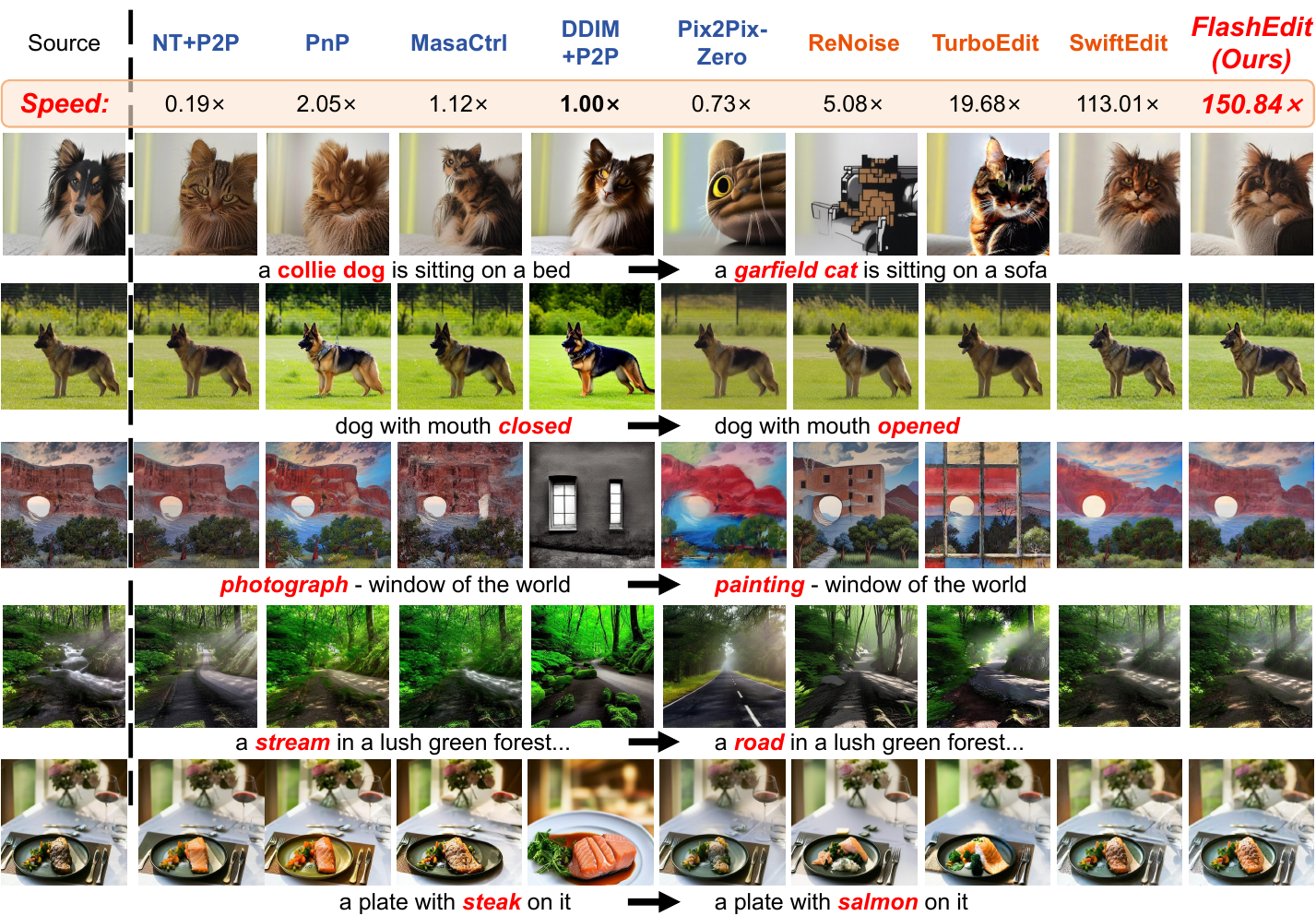}

\end{center}
\caption{\textbf{Qualitative comparison on PIE-Bench.} Each row shows one editing task, with the source image in the first column and source/target prompts listed below. FlashEdit aims to apply the requested edit while preserving regions outside the edit mask.}
    \label{fig:qualitative_comparison}
\end{figure}

\noindent\textbf{Soft Recomposition with Attention Feathering.$\quad$} During the editing pass with the target prompt $c_{tgt}$, we compute new queries, keys, and values ($Q_{tgt}, K_{tgt}, V_{tgt} \in \mathbb{R}^{S \times d_k}$). We then construct a spatially-aware, full key-value set, $K_{full}, V_{full}$, by combining the preserved-region memory with the current editable-region features:
    \begin{equation}
        K_{full}[j, :] = \begin{cases} K^*_{\mathcal{B}}[\text{rank}_{\mathcal{B}}(j), :] & \text{if } j \in \mathcal{B} \\ K_{tgt}[j, :] & \text{if } j \in \mathcal{F} \end{cases} \quad
    \end{equation}
    \begin{equation}
        V_{full}[j, :] = \begin{cases} V^*_{\mathcal{B}}[\text{rank}_{\mathcal{B}}(j), :] & \text{if } j \in \mathcal{B} \\ V_{tgt}[j, :] & \text{if } j \in \mathcal{F} \end{cases}
    \end{equation}
    where $\text{rank}_{\mathcal{B}}(j)$ maintains positional alignment.

    To mitigate boundary artifacts and encourage a smooth blend, we replace the hard-masking of the original ``foreground core'' approach with an \textbf{Attention Feathering} mechanism. We first create a soft alpha mask $A \in [0, 1]^S$ by applying a Gaussian blur to the binary mask $M$:
    \begin{equation}
        A = \text{GaussianBlur}(M, \text{kernel\_size}=\sigma)
    \end{equation}
    This alpha mask creates a ``transition zone''. This zone softly connects the editable region, where the mask value is close to one, and the preserved region, where the mask value is close to zero.

    Next, we compute the full attention output, $H_{edit} \in \mathbb{R}^{S \times d_k}$, where all queries $Q_{tgt}$ attend to the recomposed key-value set $(K_{full}, V_{full})$:
    \begin{equation}
        H_{edit} = \text{softmax}\left(\frac{Q_{tgt} K_{full}^T}{\sqrt{d_k}}\right) V_{full}.
        \label{eq:expanded_attn}
    \end{equation}
    The final sparse output matrix $H$ is then obtained by $H_{edit}$ with our soft alpha mask $A$:
    \begin{equation}
        H = A \odot H_{edit}
    \end{equation}
    where $\odot$ denotes element-wise multiplication.

\begin{table}[t]
\centering
\small
\caption{\textbf{Quantitative comparison on PIE-Bench.} We report preservation metrics on regions outside the edit mask and CLIP similarity for whole-image and edited-region semantic alignment.}
\label{tab:quality_comparison}
\setlength{\tabcolsep}{5pt} 
\renewcommand{\arraystretch}{1.1} 
\begin{tabular}{@{}ll cccc cc@{}}
\toprule\toprule
\multicolumn{2}{c}{\textbf{Method}} & \multicolumn{4}{c}{\textbf{\makecell{Background Preservation}}} & \multicolumn{2}{c}{\textbf{CLIP Similarity}} \\
\cmidrule(r){1-2} \cmidrule(lr){3-6} \cmidrule(l){7-8}
\textbf{Inverse} & \textbf{Editing} & \textbf{PSNR} $\uparrow$ & \textbf{LPIPS}$_{_{\times 10^3}}$ $\downarrow$ & \textbf{MSE}$_{_{\times 10^4}}$ $\downarrow$ & \textbf{SSIM}$_{_{\times 10^2}}$ $\uparrow$ & \textbf{Whole} $\uparrow$ & \textbf{Edited} $\uparrow$ \\
\midrule
DDIM & P2P & 17.87 & 208.80 & 219.88 & 71.14 & 25.01 & 22.44 \\
NT-Inv & P2P & 27.03 & 60.67 & 35.86 & 84.11 & 24.75 & 21.86 \\
DDIM & MasaCtrl & 22.17 & 106.62 & 86.97 & 79.67 & 23.96 & 21.16 \\
Direct Inversion & MasaCtrl & 22.64 & 87.94 & 81.09 & 81.33 & 24.38 & 21.35 \\
DDIM & P2P-Zero & 20.44 & 172.22 & 144.12 & 74.67 & 22.80 & 20.54 \\
Direct Inversion & P2P-Zero & 21.53 & 138.98 & 127.32 & 77.05 & 23.31 & 21.05 \\
DDIM & PnP & 22.28 & 113.46 & 83.64 & 79.05 & 25.41 & 22.55 \\
Direct Inversion & PnP & 22.46 & 106.06 & 80.45 & 79.68 & 25.41 & 22.62 \\
\midrule
\multicolumn{2}{l}{{ReNoise(SDXL)}} & 20.85 & 176.84 & 51.78 & 72.44 & 24.41 & 21.88 \\
\multicolumn{2}{l}{{TurboEdit}} & 22.51 & 107.27 & 9.32 & 80.09 & 25.49 & 21.82 \\
\multicolumn{2}{l}{{SwiftEdit}}  & 23.31 & 71.54 & 6.18 & 82.25 & 25.56 & 21.91 \\
\midrule
\rowcolor{orange!20}\multicolumn{2}{l}{\textbf{FlashEdit}} & 25.33 & 62.25 & 4.28 & 83.32 & 25.51 & 22.29 \\
\rowcolor{orange!20}\multicolumn{2}{l}{\textbf{FlashEdit(w/ GT masks)}} & 25.28 & 62.43 & 4.33 & 83.02 & 25.58 & 22.32 \\
\bottomrule\bottomrule
\end{tabular}
\end{table}

\subsection{Sparsified Spatial Cross-Attention}

\noindent\textbf{Challenge: Semantic Entanglement in Image Editing.$\quad$}
A key challenge in precise image editing is \textit{semantic entanglement}, where textual attributes are not cleanly bound to their intended objects. As demonstrated in Figure~\ref{fig:method3}, when changing ``a cat with yellow eyes'' to ``a cat with green eyes,'' standard cross-attention often fails to isolate the edit region. This failure stems from the inherent \textit{all-to-all} nature of the softmax function in vanilla cross-attention layers. Since every pixel is forced to compute a probability distribution across all text tokens, a highly salient but spatially irrelevant token (e.g., ``green'') can dominate the attention map. This leads to two critical artifacts: \textit{edit attenuation}, where the intended change is suppressed by structural tokens, and \textit{attribute leakage}, where the new attribute bleeds onto surrounding unedited regions (e.g., an unnatural green tint on the cat's fur). The root cause is that the softmax normalization allows unrelated semantic signals to interfere with each other spatially.

\noindent\textbf{Motivation: Pre-emptive Semantic Disentanglement.$\quad$}
Based on this diagnosis, we contend that semantic concepts must be disentangled \textit{before} the non-linear softmax allows them to compete globally. Our motivation is to implement a \textbf{pre-emptive disentanglement} strategy that restricts the receptive field of each pixel in the textual domain. By forcing the model to attend only to a clean, task-relevant subset of tokens for a given spatial region, we can eliminate attribute bleeding at its source. We introduce \textbf{Sparsified Spatial Cross-Attention (SSCA)}, which treats text attention not as a global competition, but as a localized, sparse retrieval process that preserves the high-frequency semantic boundaries of the edit.

\noindent\textbf{Proposed Method: Sparse-to-Dense Integration.$\quad$}
As illustrated in Figure~\ref{fig:method3}, the SSCA mechanism redefines text-guided guidance through a three-stage pipeline of identification, pruning, and structural integration.

\noindent\textbf{Identifying Key Semantic Tokens.$\quad$} 
Rather than processing the entire prompt $y$, we first localize the most significant textual cues for the edit region defined by mask $M$. We compute the aggregate similarity between the editable-region image queries $Q_{l,\mathcal{F}}$ and the full set of text keys $K_y$. Specifically, for each text token $j$, we compute a relevance score $R_j = \sum_{i \in \mathcal{F}} (Q_{l,i} \cdot K_{y,j}^T) / \sqrt{d}$. We then select the \textit{top-k} text key-value pairs with the highest scores, denoted as $(K_y^k, V_y^k)$. We use $k=10$ in all experiments and perform the selection independently at each applied cross-attention layer. This operation acts as a semantic filter, reducing the influence of irrelevant tokens (e.g., "background", "sky") that might otherwise dilute the masked edit.

\noindent\textbf{Computing Focused Sparse Attention Signal.$\quad$} 
With the pruned semantic subset, we compute a focused attention result $A_{\text{sparse}}$ exclusively for the editable-region queries $Q_{l,\mathcal{F}}$. 

The sparse attention for an editable-region pixel $i \in \mathcal{F}$ is computed as:
\begin{equation}
    A_{\text{sparse}, i} = \text{softmax}\left(\frac{Q_{l,i} (K_y^k)^T}{\sqrt{d}}\right) V_y^k.
\end{equation}
By restricting the denominator of the softmax to only $k$ relevant tokens, we concentrate the semantic energy of the target attribute (e.g., ``green'') on the intended spatial coordinates and reduce leakage into the global feature mean.

\noindent\textbf{Structural Integration via Zero-Masking.$\quad$} 
To restrict target-prompt influence to the editable region, we scatter the sparse results into a zero-initialized full-size attention matrix $A_{\text{SSCA}} \in \mathbb{R}^{S \times d}$:
\begin{equation}
    A_{\text{SSCA}}[i, :] = 
    \begin{cases} 
        A_{\text{sparse}}[\text{rank}_{\mathcal{F}}(i), :] & \text{if } i \in \mathcal{F} \\
        \mathbf{0} & \text{if } i \notin \mathcal{F}
    \end{cases}.
\end{equation}
This structural zero-masking encourages a clean decoupling between editable-region semantics and preserved-region textures. The final output is integrated into the model through the standard residual connection, resulting in an attribute-faithful edit that remains spatially concentrated in the masked region.

\begin{table}[t]
\centering
\small
\caption{\textbf{Ablation study on core components.} We evaluate each module within the same one-step editing pipeline. COSI serves as the inversion backbone, while BG-Shield and SSCA add spatial and semantic controls on top of this backbone.}
\label{tab:ablation_components_full}
\setlength{\tabcolsep}{6pt} 
\renewcommand{\arraystretch}{1.2} 

\begin{tabular}{@{}ccc cccc cc@{}}
\toprule
\multicolumn{3}{c}{\textbf{Components}} & \multicolumn{4}{c}{\textbf{Background Preservation}} & \multicolumn{2}{c}{\textbf{CLIP Similarity}} \\
\cmidrule(lr){1-3} \cmidrule(lr){4-7} \cmidrule(l){8-9}
\textbf{COSI} & \textbf{BG-Shield} & \textbf{SSCA} & \textbf{PSNR}$\uparrow$ & \textbf{LPIPS}$_{\times 10^{3}}$$\downarrow$ & \textbf{MSE}$_{\times 10^{4}}$$\downarrow$ & \textbf{SSIM}$_{\times 10^{2}}$$\uparrow$ & \textbf{Whole}$\uparrow$ & \textbf{Edited}$\uparrow$ \\
\midrule

\checkmark & - & - & 23.45 & 71.38 & 6.10 & 82.25 & 24.48 & 21.28 \\ 

\checkmark & \checkmark & - & 24.69 & 75.28 & 4.96 & 81.95 & 24.80 & 21.28 \\ 

\checkmark & \checkmark & \checkmark & 25.33 & 62.25 & 4.28 & 83.32 & 25.51 & 22.29 \\ 
\bottomrule
\end{tabular}
\end{table}

\section{Experiment}
\label{sec:experiment}

\subsection{Experimental Settings}

\noindent\textbf{Implementation Details.$\quad$}
We implement our framework and architecture in PyTorch inspired by~\cite{wang2025osdface,nguyen2025swiftedit,chen2025dove}, using the Adam optimizer~\cite{adam}. We evaluate our method on the PieBench benchmark~\cite{pnp}, which features 700 samples across 10 editing types. Following the default PieBench evaluation protocol, all images and masks are evaluated at $512\times512$ resolution with batch size 1. We report metrics along two primary axes. As for \textbf{Background Preservation}, We compute PSNR~\cite{PSNR}, LPIPS~\cite{lpips}, MSE and SSIM~\cite{ssim} on the unedited regions to measure fidelity to the source image. As for \textbf{Semantic Alignment}, We report CLIP-Whole~\cite{clip} for prompt-image alignment and CLIP-Edited~\cite{clip} for alignment within masked edit regions. Experiments were conducted on a NVIDIA A6000 GPU.

\noindent\textbf{Baselines.$\quad$}
We compare our method against representative baselines under the same PIE-Bench protocol. For \textbf{multi-step} methods, we evaluate Prompt-to-Prompt (P2P)~\cite{p2p}, MasaCtrl~\cite{masactrl}, Pix2Pix-Zero~\cite{pix2pixZero}, and Plug-and-Play (PnP)~\cite{pnp}, paired with inversion techniques like DDIM~\cite{ddim}, Null-text Inversion (NT-Inv)~\cite{NT-Inv}, and Direct Inversion~\cite{directInversion}. For \textbf{few-steps} and \textbf{one-step} methods, we compare against Renoise~\cite{renoise}, TurboEdit~\cite{turboedit} and SwiftEdit~\cite{nguyen2025swiftedit}. Our implementation targets the U-Net-based latent diffusion setting used by these inversion-based editing baselines; adapting the same control principles to DiT or rectified-flow backbones requires architecture-specific design and is outside the scope of this work.

\subsection{Quantitative Analysis}
\begin{wraptable}{r}{0.50\textwidth}
\vspace{-8pt}
\centering
\scriptsize
\caption{\textbf{Efficiency comparison.} We report the number of denoising steps and relative speedup for each method under the same evaluation setting.}
\label{tab:efficiency_detailed_merged}
\vspace{-4pt}
\setlength{\tabcolsep}{2.4pt}
\renewcommand{\arraystretch}{1.1}
\begin{tabular}{@{}llcc@{}}
\toprule\toprule
\multicolumn{2}{c}{\textbf{Method}} & \multirow{2}{*}{\makecell{\textbf{Denoising}\\\textbf{Steps}}} & \multirow{2}{*}{\textbf{Speedup}} \\
\cmidrule(r){1-2}
\textbf{Inverse} & \textbf{Editing} & & \\
\midrule
DDIM & P2P & \multirow{8}{*}{Multi-steps} & \textbf{1.00$\times$} \\
NT-Inv & P2P & & 0.19$\times$ \\
DDIM & MasaCtrl & & 1.12$\times$ \\
Direct Inversion & MasaCtrl & & 0.88$\times$ \\
DDIM & P2P-Zero & & 0.73$\times$ \\
Direct Inversion & P2P-Zero & & 0.73$\times$ \\
DDIM & PnP & & 2.06$\times$ \\
Direct Inversion & PnP & & 2.03$\times$ \\
\midrule
\multicolumn{2}{l}{ReNoise(SDXL)} & \multirow{2}{*}{Few-steps} & 5.08$\times$ \\
\multicolumn{2}{l}{TurboEdit} & & 19.68$\times$ \\
\midrule
\multicolumn{2}{l}{SwiftEdit} & \multirow{2}{*}{One-step}& 113.01$\times$ \\
\multicolumn{2}{l}{\cellcolor{orange!20}\textbf{FlashEdit(Ours)}} & & \cellcolor{orange!20}\textbf{150.84$\times$} \\
\bottomrule\bottomrule
\end{tabular}
\vspace{-8pt}
\end{wraptable}

As shown in \Cref{tab:quality_comparison}, FlashEdit provides a strong preservation-efficiency trade-off for accelerated localized editing. It achieves competitive quality compared with representative \textbf{few-step} methods like ReNoise~\cite{renoise} and TurboEdit~\cite{turboedit}, while maintaining background preservation comparable to strong but prohibitively slow \textbf{multi-step} inversion-based methods. This quality is delivered with an efficiency gain of over \textbf{150$\times$} against DDIM+P2P (\Cref{tab:efficiency_detailed_merged}). Furthermore, an experiment using ground-truth (GT) masks reveals a negligible performance difference, suggesting that the self-guided masks used in our PIE-Bench evaluation are close to the GT masks for this protocol.

\subsection{Qualitative Analysis}
Visual comparisons in Figure~\ref{fig:qualitative_comparison} reinforce our quantitative findings. The outputs from FlashEdit exhibit high semantic fidelity to the target prompt while maintaining strong preservation of unedited regions, reducing the ``bleeding'' artifacts common in other methods. In contrast, other baselines often display noticeable quality degradation or changes outside the masked edit region. FlashEdit therefore provides a favorable balance between visual quality and real-time performance under the standard PIE-Bench setting.

\subsection{Ablation Studies}
To validate the contribution of each component in our framework, we conduct a comprehensive ablation study, with the results presented in \Cref{tab:ablation_components_full}. We emphasize that the COSI-only row should be interpreted as an \emph{inversion backbone} rather than a complete editing system: its goal is to provide a stable, one-step latent initialization with strong reconstruction and consistency outside the edit mask, while spatial isolation and semantic disentanglement are intentionally handled by the subsequent modules. Under this controlled setting, COSI already provides a competitive one-step starting point, but it is not expected to maximize edited-region semantic alignment by itself.

Integrating \textbf{BG-Shield} brings a marked improvement across preservation metrics, supporting its effectiveness in isolating features outside the edit mask once COSI has provided a stable latent. The final addition of \textbf{SSCA} further boosts metrics. It substantially enhances semantic alignment, evidenced by a large increase in the CLIP-Edited score, which validates our pre-softmax token pruning strategy. SSCA also improves reconstruction quality, suggesting a synergistic effect where cleaner textual guidance benefits the entire process. This demonstrates that all three components work in concert to achieve the final preservation-efficiency trade-off of \textbf{FlashEdit}, rather than any single module being solely responsible for all quality gains.

\section{Conclusion}
This paper introduces \textbf{FlashEdit}, a real-time localized text-guided image editing framework for inversion-based diffusion editing. We show that a strong speed-quality trade-off can be achieved under the standard PIE-Bench protocol with a holistic, multi-level control strategy. Our approach improves geometric stability through the \textbf{COSI} pipeline for manifold-aware inversion, preserves regions outside the edit mask with the \textbf{BG-Shield} mechanism, and improves fine-grained semantic precision via \textbf{SSCA}.

\newpage
\bibliographystyle{unsrtnat}
\bibliography{example_paper}

\end{document}